\documentclass[letterpaper]{article} 
\usepackage{aaai2026}  
\usepackage{times}  
\usepackage{helvet}  
\usepackage{courier}  
\usepackage[hyphens]{url}  
\usepackage{graphicx} 
\usepackage{natbib} 
\usepackage{caption} 
\usepackage{algorithm}
\usepackage{algorithmic}
\usepackage{booktabs}
\usepackage{amsmath,amssymb}
\usepackage{url}
\usepackage{microtype}
\usepackage{array}
\usepackage{latexsym}
\usepackage{tabularx}
\usepackage{makecell}

\usepackage{pgfplots}
\pgfplotsset{compat=1.18}

\usepackage{tikz}
\usetikzlibrary{arrows.meta, positioning}

\usepackage[utf8]{inputenc}
\usepackage[T1]{fontenc}
\usepackage{textcomp}
\DeclareUnicodeCharacter{2212}{\textminus}

\usepackage{enumitem}
\usepackage{inconsolata}

\usepackage{newfloat}
\usepackage{listings}

\DeclareCaptionStyle{ruled}{
  labelfont=normalfont,
  labelsep=colon,
  strut=off
}

\floatstyle{ruled}
\newfloat{listing}{tb}{lst}{}
\floatname{listing}{Listing}

\title{Trace Integrity for LLM Data Agents: Auditing Structured-Data Computation Beyond Answer Accuracy in Real-World Systems}
\author {
    Srimonti Dutta\textsuperscript{\rm 1},
    Akshata Kishore Moharir\textsuperscript{\rm 2}
}

\affiliations {
    \textsuperscript{\rm 1}Indian Institute of Technology Guwahati\\
    \textsuperscript{\rm 2}University of Maryland\\
    srimonti.dutta@gmail.com, akshatan@terpmail.umd.edu
}

\nocopyright

\begin{document}

\maketitle

\begin{abstract}
Large language model (LLM) data agents answer structured-data questions by generating SQL, invoking tools, executing queries, and summarizing results. In operational settings, reliability depends on both the reported answer and the computation used to produce it, because errors may remain undetected when the resulting answer happens to match an expected value. We refer to the mismatch between natural-language intent and the operator-level computation required to satisfy that intent as the Structure Gap, and define Trace Integrity as a reliability criterion for assessing whether the computation supporting an answer is explicit, executable, schema-valid, operator-faithful, replayable, answer-consistent, and auditable.

We operationalize Trace Integrity through execution contracts that record the schema elements, operator plan, assumptions, executable query, verification status, and final-answer linkage associated with a request. We also define the Correct Answer / Invalid Trace (CAIT) Rate, which measures the proportion of correct answers whose recorded computations fail trace-integrity checks. Together, these mechanisms provide a way to evaluate whether answer-level success is supported by a valid and inspectable computation. We study this distinction on 100 BIRD Mini-Dev text-to-SQL examples using three prompting conditions: Direct SQL, Operation Summary + SQL, and Contract-First SQL. The three conditions achieve answer accuracies of 20.0\%, 22.0\%, and 24.0\%, Trace Integrity Pass Rates of 39.0\%, 43.0\%, and 40.0\%, and CAIT Rates of 55.0\%, 59.1\%, and 45.8\%, respectively. The results show that answer accuracy, trace validity, and the frequency of correct answers supported by invalid traces capture different aspects of system behavior. Trace-level evaluation therefore complements conventional answer-based evaluation by providing evidence about the computational support, inspectability, and reproducibility of LLM data-agent outputs in real-world systems.
\end{abstract}

\begin{links}
    \link{Code}{https://github.com/srimontidutta/trace-integrity}
\end{links}

\section{Introduction}

LLM data agents are becoming interfaces to operational data. They inspect schemas, generate SQL, call tools, execute queries, summarize results, and present conclusions to analysts, operators, clinicians, auditors, and customers. This makes databases, spreadsheets, logs, records, and enterprise analytics systems easier to query, but it also changes the reliability problem. The question is no longer only whether the system returns the expected value. A data agent may return a plausible or benchmark-correct answer while the computation behind that answer applies the wrong filter, join, aggregation, time window, grouping key, or schema binding.

Consider a business analyst asking, \emph{Which region had the highest average revenue per active customer last quarter, excluding trial accounts?} A system can return a confident region name while silently computing total revenue instead of average revenue, including trial accounts, grouping by sales office rather than customer region, using a calendar quarter instead of the fiscal quarter, or joining customer and revenue records on a brittle name field rather than a stable identifier. Answer-only evaluation cannot distinguish a faithful computation from an accidental success. A deployment team needs to know not only what answer was produced, but what computation was actually performed. This is a deployment problem in addition to being a benchmark problem. In operational workflows, the answer is often consumed by someone who cannot reconstruct the query: a BI user, compliance reviewer, data engineer, or domain expert. These users require an inspectable artifact that records the schema elements, filters, joins, aggregations, assumptions, and execution results that support the answer.

This paper argues that real-world LLM data agents need a reliability criterion that treats the computation trace as a first-class object. We call this criterion \emph{Trace Integrity}. A trace has integrity when the recorded computation behind an answer is explicit, executable, schema-valid, operator-faithful, replayable, answer-consistent, and auditable. The claim is narrower than general trustworthiness, as Trace Integrity does not certify that the underlying data are complete, fair, or appropriate for a downstream decision, instead it addresses a specific deployment risk: answer correctness is not sufficient evidence that the system performed the computation requested by the user.

The need for trace-level evaluation follows from a broader Structure Gap. Language models express reasoning as token sequences, while structured-data tasks require operator-level programs with well-defined semantics. A valid answer may depend on projection, filtering, joining, grouping, aggregation, sorting, ranking, temporal constraints, and schema binding. Text-to-SQL and table-reasoning benchmarks expose this pressure because success depends on mapping a natural-language request to an executable computation, not merely producing a plausible explanation \citep{zhong2017seq2sql,yu2018spider,pasupat2015compositional,chen2019tabfact,chen2020hybridqa}. Recent LLM-driven Text-to-SQL systems further show this problem is critical in business intelligence and inventory-management settings, where natural-language access to operational databases is important as non-specialists need to query structured data \citep{zhu2023talk}. 

Existing methods address adjacent parts of the problem but do not make Trace Integrity a first-class evaluation target. Chain-of-thought prompting can elicit intermediate text, but natural-language rationales are not executable and may not faithfully describe the computation that produced the answer \citep{wei2022chain,kojima2022large,lanham2023measuring}. Program-aided and tool-using methods improve task performance by delegating computation to external systems \citep{gao2023pal,chen2023program,yao2023react,schick2023toolformer}, yet execution alone does not show if the generated program matches the user's intended computation. An SQL query can run successfully and still answer the wrong question. Semantic-parsing work has similarly shown that correct denotation can arise from the wrong program, motivating text-to-SQL evaluation across multiple database instances \citep{zhong2020semantic}. Trace Integrity addresses a complementary deployment question by retaining and auditing the computation that supported a particular agent response.

The central hidden-failure case is a correct answer with an invalid trace. Such an output receives credit under answer-only evaluation, yet it is not a reliable data-agent response as the recorded computation does not support the answer. We call this case Correct Answer / Invalid Trace and measure it with CAIT Rate. CAIT asks: among the answers a system gets right, what fraction are backed by invalid computations? A high CAIT Rate means that answer-only evaluation is counting computationally unsupported outputs as successes.

This work develops the vocabulary, artifact, and measurement protocol for that hidden quadrant. The Structure Gap names the deployment failure mode in which natural-language reasoning fails to specify operator-level computation, Trace Integrity gives the reliability criterion, execution contracts make the criterion operational, the Isolation Principle provides a default planning discipline, and CAIT Rate turns correct-answer/invalid-trace failures into a metric that can be reported across systems and datasets.

We include a proof-of-concept to show that this deployment risk appears in measurable system behavior. On 100 BIRD Mini-Dev examples, we compare Direct SQL, Operation Summary + SQL, and Contract-First SQL. The three methods achieve answer accuracies of 20.0\%, 22.0\%, and 24.0\%, while their Trace Integrity Pass Rates are 39.0\%, 43.0\%, and 40.0\%. Their CAIT Rates remain high at 55.0\%, 59.1\%, and 45.8\%. These results show that answer accuracy, trace validity, and silent-failure risk are distinct signals that deployment teams should monitor together. The intended contribution is a practical evaluation frame for LLM data agents: if LLM data agents are to serve as reliable interfaces to operational data, they should be evaluated not only by what answer they return, but by whether the computation behind that answer can be inspected, replayed, and trusted.

\section{The Trace Integrity Problem}

\subsection{Operator Commitments and the Structure Gap}

Structured-data questions require operational commitments: which records are included, which entities are aligned, which measure is computed, which grouping defines the comparison, which temporal window is used, and which ordering determines a superlative. If a user asks for average revenue per active customer, the computation must define both numerator and denominator; if they ask for last quarter, the computation must resolve a time interval; if they ask for a region-level result, the computation must bind the intended region field rather than a superficially similar location attribute.

We define the \emph{Structure Gap} as the mismatch between natural-language reasoning and operator-level computation. Natural language expresses intent, but it does not enforce the discrete operations that structured-data answers require: filtering, joining, grouping, aggregation, ranking, temporal restriction, and schema binding. A model can state the intended operation plausibly while omitting a join, changing an average into a sum, applying a filter to the wrong table, or grouping by a field with similar surface form but different semantics. In a deployed data agent, such an error is a reliability failure because the answer is no longer supported by the computation. This framing is representational rather than a claim that language models lack general reasoning ability. A valid structured-data answer must bind a user request to an executable computation. Text-to-SQL tasks make this visible because the target output is an operator-level representation \citep{zhong2017seq2sql,yu2018spider}. Table reasoning and fact verification tasks show the same pressure in less strictly relational settings \citep{pasupat2015compositional,chen2019tabfact,chen2020hybridqa}.

\subsection{Silent Failures in Operational Data Workflows}

Silent structured-data failures are dangerous because they look like normal completions. A SQL-generating agent may produce a query that executes cleanly but groups by the wrong key. A tool-using agent may call the correct database but omit a required exclusion filter. A rationale may say that trial accounts were excluded while the executed query includes them. In each case, the interface returns an answer and gives the user little reason to suspect that the computation has drifted from the request. 
These failures appear in ordinary data-agent workflows. In business intelligence, a model may report total revenue when the decision depends on revenue per customer. In compliance reporting, it may count all transactions rather than transactions in the regulated jurisdiction. In clinical analytics, it may define a cohort using visit dates rather than diagnosis dates. The deployment consequence is that a final answer is not enough of a reliability artifact: the computation must be available for inspection. 

\begin{figure}[t]
\centering
\small
\fbox{\begin{minipage}{0.94\linewidth}
\centering
User query $\rightarrow$ schema and policy context \\
$\rightarrow$ execution contract $\rightarrow$ validation \\
$\rightarrow$ deterministic execution $\rightarrow$ final answer \\
$\rightarrow$ auditable trace store
\end{minipage}}
\caption{Contract-based LLM data-agent lifecycle.}
\label{fig:lifecycle}
\end{figure}

\subsection{The CAIT Failure Mode}

Answer accuracy collapses several reliability states into one outcome. A correct answer with a valid trace is a faithful success. An incorrect answer with an invalid trace is a visible structural failure. An incorrect answer with a valid trace may reflect a data issue, an execution issue, an ambiguity in the user request, or a mismatch between the benchmark reference and another defensible interpretation. The most deployment-relevant hidden case is a correct answer with an invalid trace. This output receives credit under answer-only evaluation even though the recorded computation does not support the answer.

We call this hidden-success case \emph{Correct Answer / Invalid Trace} and measure it with CAIT Rate. CAIT asks: among the answers a system gets right, what fraction are backed by invalid computations? Formally,
\[
\mathrm{CAIT}
=
\frac{N_{\mathrm{correct}\cap\mathrm{invalid}}}
     {N_{\mathrm{correct}}},
\]
where $N_{\mathrm{correct}\cap\mathrm{invalid}}$ is the number of examples whose final answer is correct but whose trace fails integrity, and $N_{\mathrm{correct}}$ is the total number of correct answers.

A high CAIT Rate means that answer-only evaluation is overstating reliability by counting computationally unsupported outputs as successes. This is important because an invalid computation may still produce the expected answer for a particular database state. A missing filter, an incorrect grouping key, a wrong join path, or an inappropriate aggregation can occasionally yield the same result as the intended computation. Such agreement is fragile: the same query may fail when the data distribution changes, when the database is updated, or when a closely related question is asked.

CAIT therefore captures a failure mode that is largely invisible at the answer level. It is not intended to replace answer accuracy or execution success, but to complement them by identifying cases in which a seemingly successful output lacks computational support. Answer accuracy measures whether the reported result matches the reference, while execution success establishes whether the generated query or program runs. Neither measurement determines whether the executed computation faithfully represents the user's request.

This separation is especially important in deployed data-agent workflows, where users may rely on the reported answer without inspecting the underlying SQL or tool trace. A query can execute successfully, return a plausible result, and even match the expected answer while still applying the wrong filter, join, aggregation, grouping key, time window, or schema binding. Trace Integrity addresses the corresponding operational question: whether the recorded computation is explicit, executable, schema-valid, operator-faithful, replayable, answer-consistent, and auditable. CAIT isolates the subset of apparently successful outputs for which that requirement is not met.

\section{Trace Integrity as an Audit Contract}

\subsection{Definition of Trace Integrity}

Trace Integrity is the property that the recorded trace behind a structured-data answer specifies the computation in a form that can be checked, replayed, and audited. The relevant trace is the system record that determines whether the answer is computationally supported: schema elements, filters, joins, grouping, aggregation, ordering, assumptions, executed query or program, and final-answer linkage. A trace passes integrity when it satisfies the dimensions in Table~\ref{tab:dimensions}. These dimensions are narrower than full semantic correctness, as they define the minimum conditions under which a data-agent answer can be treated as supported by its recorded computation. This separation is what makes CAIT Rate necessary: an answer can match the reference while the trace remains invalid.

\begin{table*}[t]
\centering
\small
\begin{tabularx}{\textwidth}{%
  >{\hsize=.55\hsize}X
  >{\hsize=1.15\hsize}X
  >{\hsize=1.30\hsize}X}
\hline
\textbf{Dimension} & \textbf{Check} & \textbf{Deployment failure exposed} \\
\hline
Explicit &
The trace records the computational commitments needed to answer the request. &
The answer gives a number but does not record the measure, grouping key, filter set, time window, or join path. \\
Executable &
The trace contains a query, program, or structured plan that can be run or deterministically checked. &
The system stores only a free-text rationale, leaving no executable artifact to validate or replay. \\
Schema-valid &
Referenced tables, columns, join keys, and fields exist in the available schema. &
The trace binds the request to \texttt{customer.region} when only \texttt{office\_region} exists. \\
Operator-faithful &
The operations preserve the user's intended computation. &
The trace uses \texttt{SUM} where the request requires a per-customer average, or groups by office rather than customer region. \\
Replayable &
Re-execution under the same data snapshot and execution environment yields the same result. &
The answer depends on hidden state, unlogged tool output, or an unspecified data snapshot. \\
Answer-consistent &
The final response follows from the executed trace. &
The query returns region A, but the natural-language response reports region B. \\
Auditable &
A reviewer or validation system can inspect the computation, assumptions, and failure mode. &
The system stores only the final answer and no structured record of how it was produced. \\
\hline
\end{tabularx}
\caption{Trace Integrity dimensions for LLM data agents.}
\label{tab:dimensions}
\end{table*}

\subsection{Execution Contracts}

An execution contract is a compact artifact that records the intended computation before or alongside execution. It binds user intent to schema elements, operator plans, assumptions, execution queries, verification status, and final answers. The contract is not a verbose explanation and it is not a substitute for deterministic execution. Its role is to make the agent's computational commitments available to validators, auditors, and downstream answer generators.

Listing~\ref{lst:contract} shows the intended level of detail. The contract does not expose hidden reasoning. It exposes the commitments that matter for computation: the tables, join key, filters, grouping field, metric, operator sequence, and verification status.

\begin{listing}[!t]
\caption{A compact execution contract recording
computational commitments for validation and audit.}
\label{lst:contract}
\begin{lstlisting}
{
  "intent": {
    "task": "top region by revenue per customer",
    "population": "active non-trial customers",
    "period": "last_quarter"
  },
  "schema": {
    "tables": ["customers", "revenue"],
    "join": {
      "left": "customers.customer_id",
      "right": "revenue.customer_id"
    },
    "filters": [
      "customers.status = 'active'",
      "customers.account_type != 'trial'",
      "revenue.date in last_quarter"
    ],
    "group_by": "customers.region",
    "metric": {
      "numerator": "SUM(revenue.amount)",
      "denominator": "COUNT(DISTINCT customers.customer_id)"
    }
  },
  "plan": [
    "filter", "join", "group_by",
    "compute_metric", "sort_desc", "top_1"
  ],
  "verification": {
    "trace_integrity": "pending"
  }
}
\end{lstlisting}
\end{listing}

A validator can check whether the referenced schema elements exist, whether the generated query contains the exclusion filter for trial accounts, whether the join uses \texttt{customer\_id}, whether the metric uses a per-customer denominator, and whether the final answer follows from the executed result.

Figure~\ref{fig:lifecycle} shows the deployment boundary created by the contract. The important aspect is not a particular agent framework, but the preserved artifact: validation can fail on schema bindings, operators, assumptions, permissions, contract-query consistency, or answer-trace consistency before the system presents a fluent answer. For an industry deployment, this is important because specific failures can be reviewed by humans and used to decide whether an answer should be shown, blocked, or escalated.

\subsection{The Isolation Principle}

The Isolation Principle states that, by default, an LLM data agent should specify its intended computation before value-level data access. The motivation is that value access can let a model complete an underspecified plan retrospectively and produce a plausible account of a computation it never committed to. Planning from the user request, schema, metadata, and policy context forces the agent to state the required operations before it sees result values that may make an invalid computation appear successful. The principle should not be applied as a rigid no-access rule. Exploratory summarization, outlier detection, profiling, deduplication, and value-dependent thresholding may require value inspection. In those cases, the contract should record why value access was necessary, what access was granted, and how it changed the plan. The deployment value of isolation is therefore that it separates specification from execution by default and makes deviations from that separation auditable.

\begin{figure}[t]
\centering
\includegraphics[width=\columnwidth]{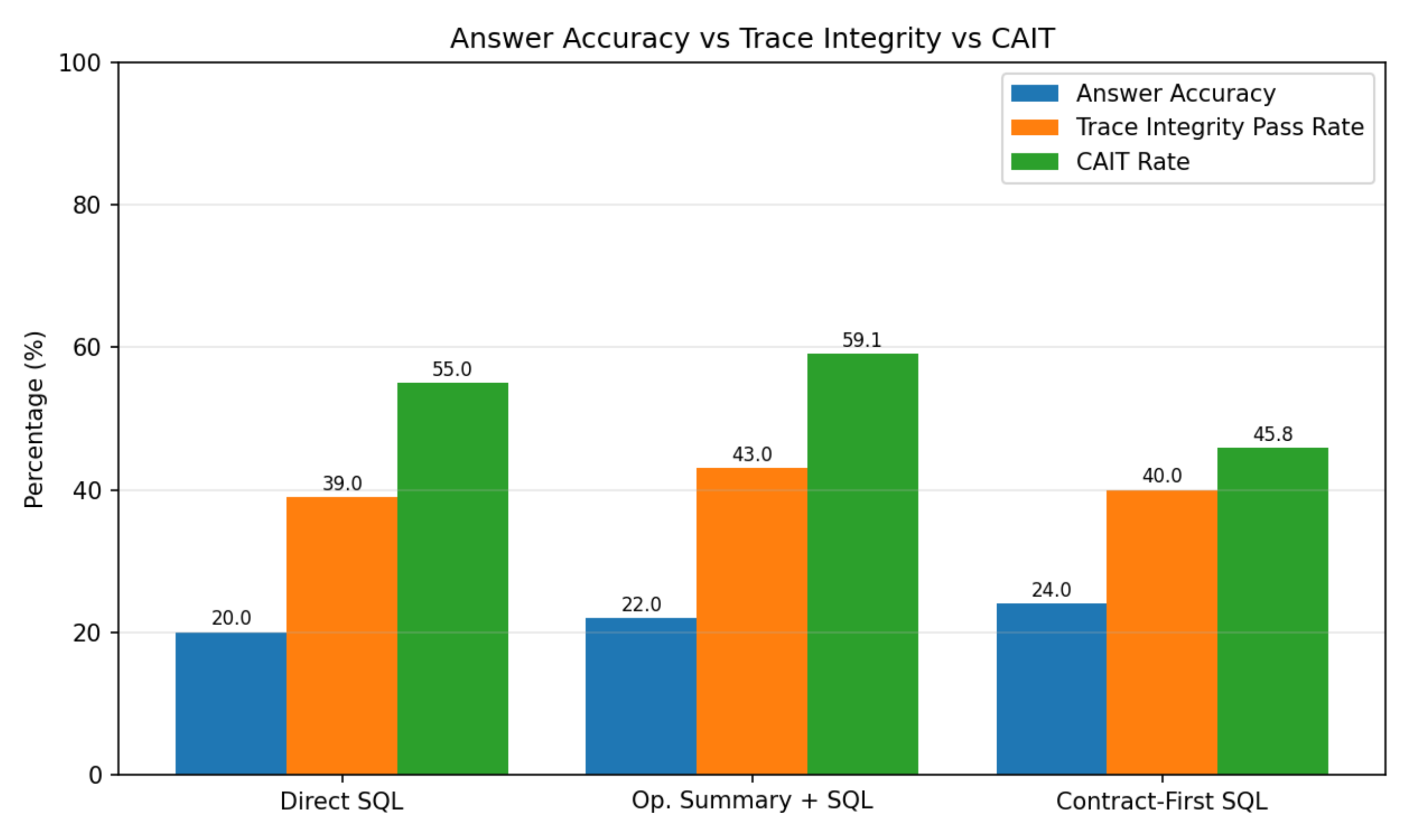}
\caption{Answer Accuracy, Trace Integrity Pass Rate, and CAIT Rate on BIRD Mini-Dev. The metrics do not move together.}
\label{fig:metric_bars}
\end{figure}

The present experiment does not isolate this principle experimentally: none of the three conditions varies value-level data access as an independent factor. Contract-First SQL tests structured commitment before query generation, while a direct test of isolation would require controlling when result values become available to the model.

\section{Empirical Demonstration: Measuring Hidden Trace Failures}

We use BIRD Mini-Dev to test whether answer correctness and trace validity diverge in a realistic database-grounded setting. BIRD is useful as it stresses larger and more heterogeneous databases than small synthetic table tasks \citep{li2023can}. This proof of concept uses BIRD Mini-Dev because the present study is concerned with data-agent behavior closer to operational analytics.

\begin{table*}[t]
\centering
\small
\begin{tabularx}{\textwidth}{Xccccc}
\toprule
\textbf{Method} & \makecell{Answer\\Accuracy} & \makecell{Execution\\Success} & \makecell{Trace Integrity Pass} & \makecell{Answer-Trace\\Consistency} & \makecell{CAIT Rate} \\
\midrule
Direct SQL & 20.0\% & 84.0\% & 39.0\% & 84.0\% & 55.0\% \\
Operation Summary + SQL & 22.0\% & 83.0\% & 43.0\% & 67.0\% & 59.1\% \\
Contract-First SQL & 24.0\% & 82.0\% & 40.0\% & 82.0\% & 45.8\% \\
\bottomrule
\end{tabularx}
\caption{Results on BIRD Mini-Dev examples. Higher is better for Answer Accuracy, Execution Success, Trace Integrity Pass Rate, and Answer-Trace Consistency. Lower is better for CAIT Rate.}
\label{tab:results}
\end{table*}

\begin{figure*}[t]
\centering
\includegraphics[width=0.82\textwidth]{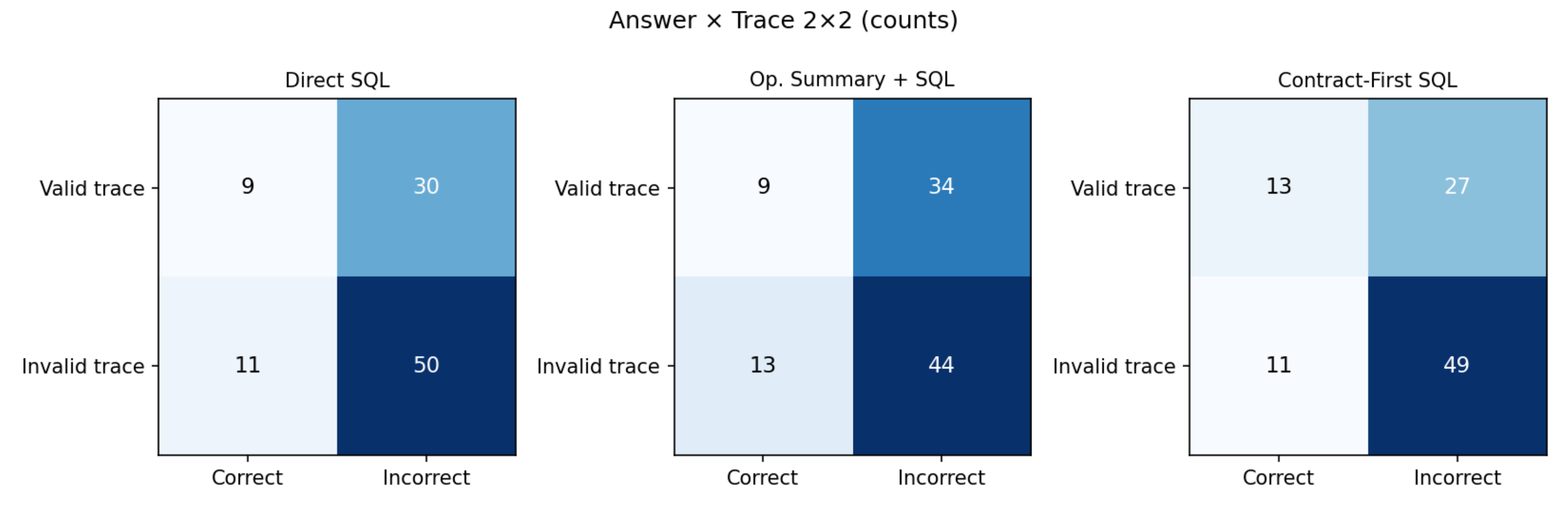}
\caption{Answer-trace quadrants by prompting condition. The lower-left cell is the CAIT case: a correct answer with an invalid trace. These cases receive credit under answer-only evaluation but are not computationally supported.}
\label{fig:answer_trace_counts}
\end{figure*}

\begin{figure*}[t]
    \centering
    \includegraphics[
        width=\textwidth,
        keepaspectratio
    ]{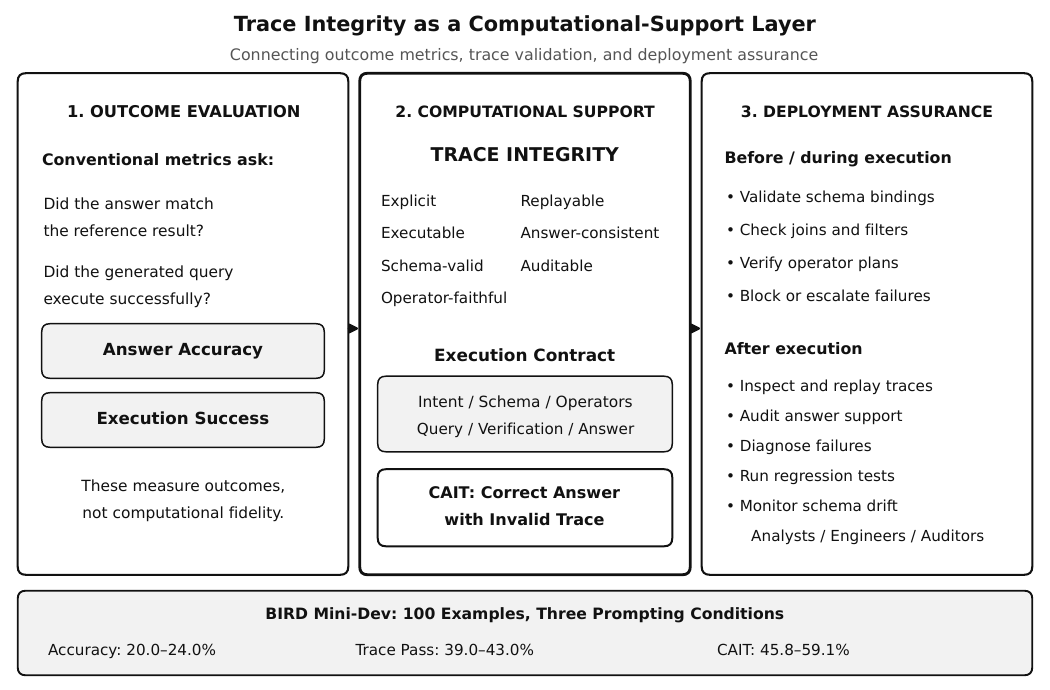}
    \caption{Trace Integrity as a computational-support layer for LLM data agents. Outcome metrics characterize reference agreement and execution behavior, whereas trace-level validation evaluates whether the recorded computation implements the requested operation and supports the reported answer. Execution contracts make the relevant computational commitments available for validation, and CAIT identifies correct answers whose traces fail integrity checks. These trace artifacts can support validation before and during execution, as well as replay, audit, failure diagnosis, regression testing, and schema-drift monitoring after execution. The BIRD Mini-Dev results illustrate that Answer Accuracy, Trace Integrity Pass Rate, and CAIT Rate characterize different properties of system behavior.}
    \label{fig:trace-integrity-layer}
\end{figure*}

\subsection{Study Design}

The proof-of-concept uses 100 BIRD Mini-Dev examples. For each example, we execute the gold SQL to obtain the reference result, execute the generated SQL against the same database, and compare the generated trace with a normalized gold trace. The primary model is \texttt{claude-haiku-4-5} with temperature 0.0. We compare three prompting conditions. Direct SQL generates a query from the question and schema. Operation Summary + SQL writes a concise natural-language summary of the intended operations before generating SQL. Contract-First SQL emits a structured execution contract before generating SQL. All three conditions receive the same question and schema context and are evaluated with the same executor and trace validator.

\subsection{Metrics and Validation}

The evaluation reports five metrics: Answer Accuracy measures whether the generated SQL result equals the gold SQL result under row-set comparison. Execution Success measures whether the generated SQL runs. Trace Integrity Pass Rate measures whether the trace is executable, schema-valid, and free of critical operator mismatches. Answer-Trace Consistency measures whether the final answer follows from the executed trace and, when present, from the declared operation summary or execution contract. CAIT Rate measures the fraction of correct answers supported by invalid traces:
\[
\mathrm{CAIT} =
\frac{N_{\mathrm{correct}\cap\mathrm{invalid}}}
     {N_{\mathrm{correct}}}.
\]
Here, \(N_{\mathrm{correct}\cap\mathrm{invalid}}\) is the number of examples whose final answer matches the reference result but whose trace fails integrity, and \(N_{\mathrm{correct}}\) is the number of examples with correct final answers. A high CAIT Rate means that answer-only evaluation is counting computationally unsupported outputs as successes.

The proof-of-concept operationalizes the dimensions that can be checked directly from the generated SQL and recorded artifacts: executability, schema validity, operator fidelity, and answer consistency. Explicitness is represented by the presence of the required structured artifact, while replayability and auditability are properties of the retained execution record and evaluation setup rather than separately scored metrics in this experiment.
Table~\ref{tab:dimensions} therefore defines the broader deployment criterion; the reported Trace Integrity Pass Rate is its narrower experimental operationalization. 

For validation, the gold SQL is normalized into an operator-level trace that records referenced schema elements and the query operations relevant to the request, including joins, filters, grouping, aggregation, ordering, and limits. Validation compares these commitments rather than SQL surface form; equivalent rewrites that the deterministic checker cannot establish as equivalent may still be flagged. The validator favors inspectable failure rules over full semantic equivalence. A trace fails integrity when it contains a critical mismatch relative to the normalized gold trace or, for Contract-First SQL, a mismatch between the contract and the generated SQL. Critical failures include wrong or missing aggregation, missing filters, missing joins, wrong join paths, wrong grouping keys, wrong sort or limit, invalid schema references, non-executable SQL, answer-trace mismatch, and contract--SQL mismatch. The checker can over-flag semantically valid rewrites when it cannot establish equivalence. CAIT in this experiment therefore combines confirmed structural mismatches with cases whose computational support the deterministic validator cannot verify, and should not be interpreted as the prevalence of semantically wrong programs alone. Distinguishing these categories through semantic equivalence testing or independent adjudication is an important next validation step.

\subsection{Results}

Answer accuracy ranges from 20.0\% to 24.0\% across the three conditions, but many correct answers are not supported by valid traces. Direct SQL obtains 20.0\% answer accuracy, 39.0\% Trace Integrity Pass Rate, and 55.0\% CAIT Rate. Operation Summary + SQL obtains 22.0\% answer accuracy, the highest Trace Integrity Pass Rate at 43.0\%, and the highest CAIT Rate at 59.1\%. Contract-First SQL obtains the highest answer accuracy at 24.0\% and the lowest CAIT Rate at 45.8\%, but its Trace Integrity Pass Rate is 40.0\%, below Operation Summary + SQL. 
Because CAIT is conditioned on only 20, 22, and 24 correct answers, respectively, its sampling uncertainty is substantial. Exact 95\% binomial intervals are 31.5-76.9\% for Direct SQL, 36.4-79.3\% for Operation Summary + SQL, and 25.6-67.2\% for Contract-First SQL. The overlapping intervals do not support a ranking of the prompting conditions by CAIT.
The result thus serves as evidence that answer accuracy, trace validity, and silent-failure risk are distinct deployment signals. Table~\ref{tab:results} shows the main result. 
Answer-Trace Consistency is evaluated separately from execution success. Their equality for Direct SQL and Contract-First SQL is an observed result of these runs rather than a consequence of the metric definition; Operation Summary + SQL separates the two measures, with 83.0\% execution success and 67.0\% answer-trace consistency.

Figure~\ref{fig:metric_bars} visualizes the same separation. Operation Summary + SQL has the highest Trace Integrity Pass Rate, while Contract-First SQL has the lowest CAIT Rate. This is important for deployment because a system can look better under one reliability signal and worse under another. Production evaluation should therefore report answer accuracy and trace-level metrics together rather than treating answer accuracy as the only success criterion.

Figure~\ref{fig:answer_trace_counts} shows the result as answer-trace quadrants. The lower-left cell is the CAIT cell: correct answer, invalid trace. Direct SQL has 11, Operation Summary + SQL has 13, and Contract-First SQL has 11 such cases. Here, answer-only evaluation counts as successes even though the recorded computation is not valid. The matrices make the central empirical point of the paper visible: apparently successful outputs are not homogeneous.

Across 300 method-example predictions, 51 queries do not execute, or 17.0\% of all predictions. Among executable outputs, Trace Integrity Pass Rates are 46.4\% (39/84) for Direct SQL, 51.8\% (43/83) for Operation Summary + SQL, and 48.8\% (40/82) for Contract-First SQL, showing that trace failures remain common after non-executable queries are excluded.
These visible failures explain part of the trace-integrity failure mass. The more important deployment lesson is that the validator also localizes failures that answer-only evaluation hides: wrong joins or aggregations, operation summaries that diverge from generated SQL, and contract-first outputs with missing filters or contract-query mismatches. The practical value of Trace Integrity: a failure can be diagnosed as a schema-validity problem, operator mismatch, contract-query inconsistency, or answer-trace inconsistency rather than disappearing behind a fluent answer.

These results also clarify the role of trace-level evaluation in a deployed data-agent system. Answer accuracy characterizes agreement with a reference result, and execution success establishes that the generated query can run. Trace Integrity evaluates a separate property: whether the recorded computation implements the requested operation and supports the reported answer. Figure~\ref{fig:trace-integrity-layer} summarizes this relationship and connects the empirical measurements to the validation and review functions enabled by execution contracts in operational systems.

\section{Deployment Implications}

The empirical results have a direct consequence for the evaluation of LLM data agents used with operational data. Answer accuracy records whether an output agrees with a reference result, while execution success records whether a generated query or program runs. Neither measurement specifies whether the computation represented in the trace implements the requested operation. Trace Integrity adds this computational-support criterion by making schema bindings, operators, assumptions, execution artifacts, and answer linkage available for validation and review.

Figure~\ref{fig:trace-integrity-layer} summarizes the role of this additional layer. In a deployed system, an execution contract can be checked before or during execution for invalid schema references, missing filters, incorrect joins, aggregation mismatches, grouping errors, temporal constraints, and inconsistencies between the declared plan and the executable query. After execution, the preserved trace can be used to verify that the reported answer follows from the executed computation. The same record also supports replay, debugging, regression testing, compliance review, and investigation of failures caused by schema or policy changes.

CAIT is particularly relevant in this setting because it identifies outputs that appear successful under answer-based evaluation while failing trace-integrity checks. Such cases can arise when an incorrect computation happens to produce the expected value for the database state used during evaluation. The underlying error may become visible after a data update, under a related query, or during an audit of the computation. For operational analytics, this makes the trace useful beyond benchmark scoring: it provides evidence about how the answer was produced and whether the recorded computation can be relied upon under review.

The proposed framework is intended to apply beyond a particular text-to-SQL benchmark or prompting method. Structured-data agents in business intelligence, financial analysis, compliance reporting, healthcare analytics, and other data-intensive applications make the same kinds of operator-level commitments about records, joins, filters, aggregations, temporal windows, and schema elements. Trace Integrity provides a common vocabulary for evaluating those commitments, while execution contracts provide a concrete artifact that systems can retain, validate, compare, and audit. This combination makes trace-level evaluation suitable for both experimental studies and production assurance workflows. 

The accompanying repository provides the Trace Integrity specification, execution-contract schema, worked examples, and evaluation guidance for applying these ideas in data-agent assessment and regression testing.

\section{Conclusion}

This paper introduced Trace Integrity as a reliability criterion for LLM data agents that produce answers through structured computation. The framework addresses the Structure Gap between natural-language intent and the operator-level program required to answer a structured-data
request. Trace Integrity defines seven properties of a supported trace: explicitness, executability, schema validity, operator fidelity, replayability, answer consistency, and auditability. Execution contracts provide a structured representation of the corresponding computational commitments, and CAIT Rate measures the fraction of correct answers whose recorded traces fail integrity checks.

The BIRD Mini-Dev study shows why these quantities should be reported separately. Across the three prompting conditions, Answer Accuracy, Trace Integrity Pass Rate, and CAIT Rate follow different patterns, and each condition produces correct answers that are associated with invalid traces. The result establishes answer-trace divergence as a measurable property of LLM data-agent behavior rather than an edge case visible only through qualitative inspection.

Trace Integrity provides a basis for evaluating whether a reported answer is supported by an inspectable and replayable computation. This criterion complements answer-level evaluation with information that is needed when data-agent outputs are reviewed, reproduced, debugged, or used in operational settings. We expect trace-level measures such as CAIT, together with structured execution artifacts, to support broader work on evaluation, auditing, and reliability for LLM agents that act over structured data.

\section{Limitations}

This study presents a scoped proof-of-concept for measuring answer-trace divergence in LLM data agents rather than a comprehensive benchmark of model families or prompting strategies. The experiment uses 100 BIRD Mini-Dev examples and one model, Claude Haiku 4.5, under a fixed prompting and execution setup. The absolute rates should therefore be interpreted as evidence that answer correctness, trace validity, and silent-failure risk can separate in practice, not as stable rankings of particular models or prompt formats.

The trace validator uses deterministic operator-level checks. This design makes the evaluation reproducible and inspectable: trace failures can be attributed to concrete issues such as missing joins, wrong aggregations, missing filters, invalid schema references, wrong grouping keys, wrong sort or limit operations, answer-trace mismatches, and contract-SQL mismatches. At the same time, structural validation is not full semantic-equivalence checking. Some SQL rewrites that are semantically valid may be penalized when the checker cannot prove equivalence to the normalized gold trace, and BIRD Mini-Dev can admit multiple valid SQL programs for the same question. The absolute answer accuracies are low relative to systems optimized for BIRD and should not be read as competitive text-to-SQL results. The study uses a deliberately simple, fixed generation setup to test whether answer-level and trace-level judgments separate under the same evaluation conditions; it does not establish a performance baseline for BIRD.
For this reason, CAIT Rate should be read as a diagnostic signal of computational support rather than as a final semantic judgment on every individual query.

Execution failures also account for a meaningful share of the trace-integrity failure mass. These failures are part of deployed text-to-SQL reliability, but they are distinct from the silent-failure case targeted by CAIT. The strongest evidence for the paper’s central claim comes from cases where execution succeeds and answer-only evaluation still fails to reveal that the recorded computation is invalid.

\section{Ethical Considerations}

This study uses BIRD Mini-Dev, an existing text-to-SQL benchmark, and does not collect new human-subjects data. The evaluation is conducted only over benchmark questions, benchmark database schemas, generated SQL outputs, execution results, and derived trace artifacts. No private user data, interactive user logs, or newly collected personally identifying information are used in the study.

The motivation for Trace Integrity is ethical as well as technical. In structured-data settings, a fluent or benchmark-correct answer can conceal an invalid computation. If such outputs are used in real-world decision-support settings, users may place trust in answers that are not actually supported by the underlying query or data transformation. Trace-level evaluation can reduce this risk by making joins, filters, aggregations, assumptions, execution results, and answer-trace consistency available for inspection. At the same time, trace artifacts create their own governance responsibilities. In deployed systems, execution contracts and trace logs may reveal sensitive details. They should therefore be protected with the same care as other operational data artifacts. Systems that store traces should apply appropriate access controls, retention limits, redaction policies, and audit logging. Trace Integrity should not be interpreted as a reason to expose hidden model reasoning or unnecessary sensitive values; the proposed contract is intended to record computational commitments, not private chain-of-thought or unrestricted data snapshots.

Trace Integrity should also be understood as a support mechanism for responsible review rather than as a replacement for domain judgment. A valid trace can make a system’s computation easier to inspect, replay, and challenge, but it does not by itself determine whether the underlying data is sufficient for a downstream decision. In high-stakes settings, trace artifacts should therefore be used alongside existing data-governance procedures, human review, and institutional oversight.

\bibliography{aaai2026}

\end{document}